\def\year{2020}\relax
\documentclass[letterpaper]{article} 
\usepackage{aaai20}  
\usepackage{times}  
\usepackage{helvet} 
\usepackage{courier}  
\usepackage[hyphens]{url}  
\usepackage{graphicx} 
\usepackage{graphicx}  
\newtheorem{example2}{\bf Example Domain}                 
\newtheorem{execexample}{\bf Execution Example}
\usepackage{subcaption}
\usepackage{amsmath,amssymb,mathtools}

\newcommand{\stt}[1]{{\small\texttt{#1}}}

\newenvironment{s_itemize}{\begin{list}{$\bullet$}
{\setlength{\rightmargin}{0em}
\setlength{\itemsep}{0em}
\setlength{\topsep}{0em}
\setlength{\parsep}{0em}}}{\end{list}}

\title{Axiom Learning and Belief Tracing for Transparent Decision
  Making in Robotics}

\author{Tiago Mota\textsuperscript{\rm 1},\\ \Large \textbf{Mohan
    Sridharan\textsuperscript{\rm 2}}\\ %
  \textsuperscript{\rm 1}Electrical and Computer Engineering, The
  University of Auckland, New Zealand\\ %
  tmot987@aucklanduni.ac.nz\\ %
  \textsuperscript{\rm 2}School of Computer Science, University of Birmingham, United Kingdom\\
  m.sridharan@bham.ac.uk }
 
\begin{document}

\maketitle

\begin{abstract}
  A robot's ability to provide descriptions of its decisions and
  beliefs promotes effective collaboration with humans. Providing such
  transparency is particularly challenging in integrated robot systems
  that include knowledge-based reasoning methods and data-driven
  learning algorithms. Towards addressing this challenge, our
  architecture couples the complementary strengths of non-monotonic
  logical reasoning, deep learning, and decision-tree induction.
  During reasoning and learning, the architecture enables a robot to
  provide on-demand relational descriptions of its decisions, beliefs,
  and the outcomes of hypothetical actions. These capabilities are
  grounded and evaluated in the context of scene understanding tasks
  and planning tasks performed using simulated images and images from
  a physical robot manipulating tabletop objects.
\end{abstract}

\section{Introduction}
\label{sec:introduction}
\begin{figure}[htb]
  \begin{center}
    \begin{subfigure}{0.23\textwidth}
      \includegraphics[width=\textwidth]{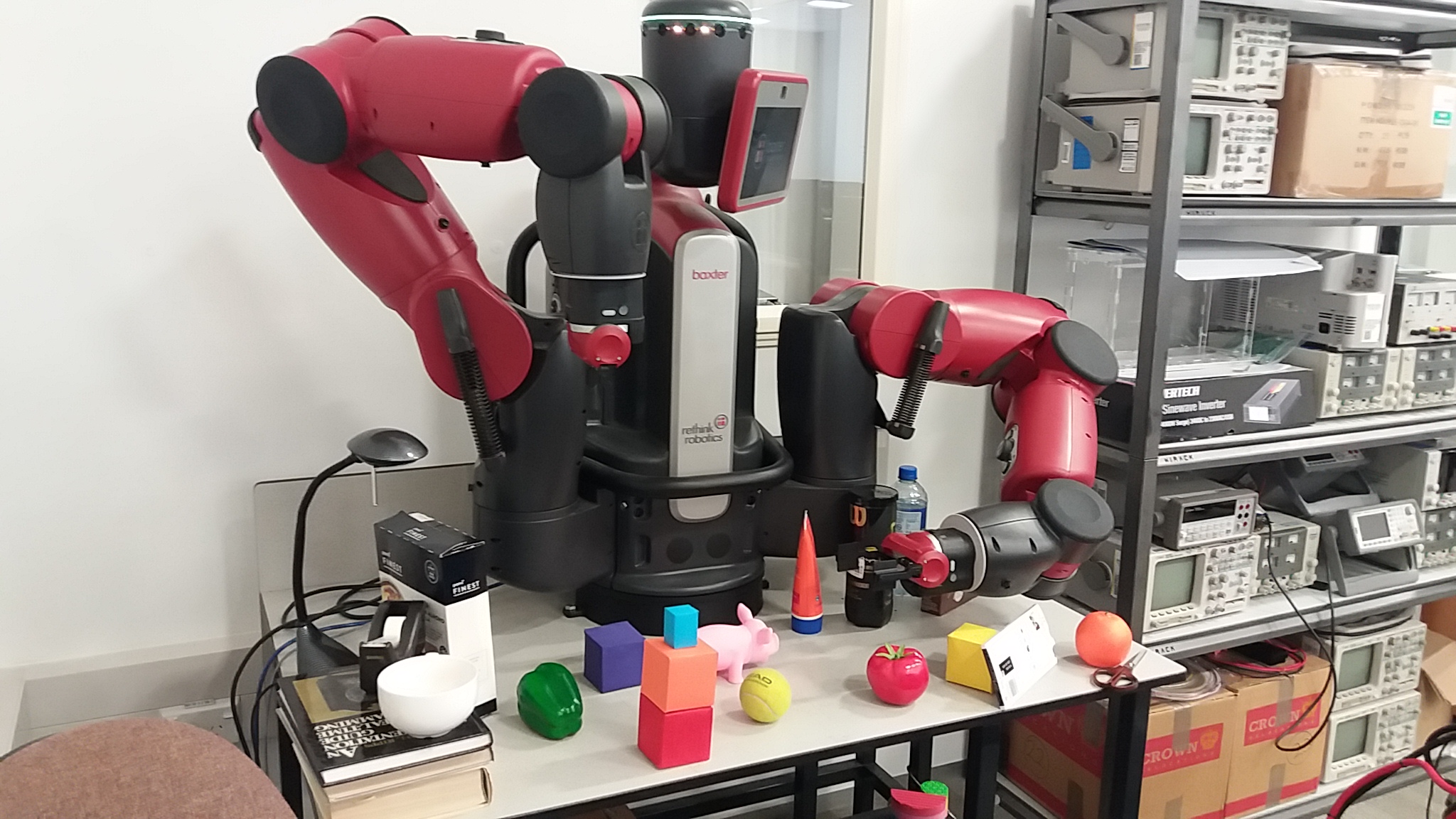}
      \caption{Test scenario.}
      \label{fig:initial}
    \end{subfigure}
    \begin{subfigure}{0.2\textwidth}
      \includegraphics[width=\textwidth]{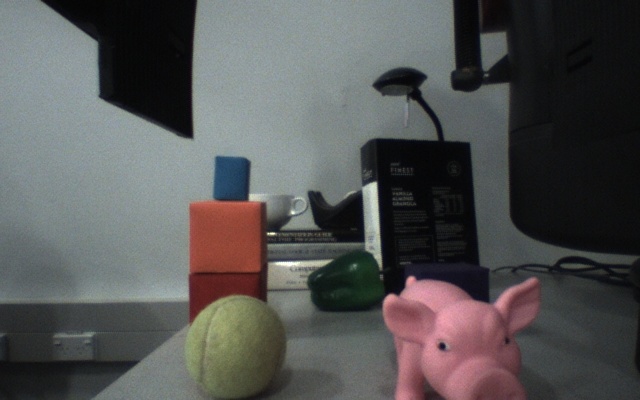}
      \caption{Robot's camera image.}
      \label{fig:left}
   \end{subfigure}
  \vspace{-1em}
  \caption{(a) Motivating scenario of a Baxter robot arranging objects
    in desired configurations on a tabletop; (b) Image from the camera
    on the robot's left gripper.}
  \label{fig:scenario-image}
  \end{center}
  \vspace{-1.75em}
\end{figure}

Consider a robot estimating the occlusion of objects and stability of
object structures while arranging objects in desired configurations on
a table; Figure~\ref{fig:initial} shows such a scene.
To perform these tasks, the robot extracts information from on-board
camera images, reasons with this information and incomplete domain
knowledge, and executes actions to achieve desired outcomes. The robot
also learns previously unknown axioms governing domain dynamics, and
provides on-demand descriptions of its decisions and beliefs.  For
instance, assume that the goal in Figure~\ref{fig:left} is to have the
yellow ball on the orange block, and that the plan is to move the blue
block on to the table before placing the ball on the orange block.
When asked to justify a plan step, e.g., ``why do you want to pick up
the blue block first?'', the robot answers ``I have to put the ball on
the orange block, and the blue block is on the orange block''; when
asked, after plan execution, ``why did you not pick up the pig?'', the
robot responds ``Because the pig is not related to the goal''.

Our work seeks to enable such on-demand \emph{explanations} of a
robot's decisions and beliefs, and hypothetical situations, in the
form of descriptions of relations between relevant objects, actions,
and domain attributes. This ``explainability'' can help improve the
underlying algorithms and establish accountability. This is
challenging to achieve with integrated robot systems that include
knowledge-based reasoning methods (e.g., for planning) and data-driven
(deep) learning algorithms (e.g., for pattern recognition). Inspired
by research in cognitive systems that indicates the benefits of
coupling different representations and reasoning
schemes~\cite{laird:book12,winston:CMHI18}, our architecture combines
the complementary strengths of knowledge-based and data-driven methods
to provide transparent decision making. It builds on our prior work
that combined non-monotonic logical reasoning and deep learning for
scene understanding in simulated images~\cite{mota:rss19}. A recent
paper described our architecture's ability to learn previously unknown
constraints and extract relevant information to construct descriptions
of decisions and beliefs~\cite{mota:eumas20}. Here, we summarize these
capabilities and describe extensions to:
\begin{itemize}
\item Incrementally acquire previously unknown action preconditions
  and effects, exploiting the interplay between representational
  choices, reasoning methods, and learning algorithms to construct
  accurate explanations.

\item Automatically trace and explain the evolution of any given
  belief from the initial beliefs by inferring the application of a
  suitable sequence of known or learned axioms.
\end{itemize}
In our implementation, non-monotonic logical reasoning is achieved
using Answer Set Prolog~\cite{balduccini:aaaisymp03}, and existing
network models are adapted for deep learning. We illustrate our
architecture's capabilities in the context of a robot (i) computing
and executing plans to arrange objects in desired configurations; and
(ii) estimating occlusion of objects and stability of object
configurations.

\section{Related Work}
\label{sec:relwork}
Early work on explanation generation drew on research in cognition,
psychology, and linguistics to characterize explanations in terms of
generality, objectivity, connectivity, relevance, and information
content~\cite{friedman:JP74}; studies with human subjects have
supported these findings~\cite{read:PSP93}. Computational methods were
also developed for explaining unexpected
outcomes~\cite{genesereth:AIJ84,dekleer:AIJ87}.

There is much interest in understanding the operation of AI and
machine learning methods, and making automation more
acceptable~\cite{miller:AIJ19}. Existing work on \textit{explainable
  AI/planning} can be broadly categorized into two groups. Methods in
one group modify or transform learned models or reasoning systems to
make their decisions more interpretable, e.g., by tracing decisions to
inputs~\cite{koh:icml17}, learning equivalent interpretable models of
any classifier~\cite{ribeiro:kdd16}, or biasing a planning system
towards making decisions easier for humans to
understand~\cite{zhang:icra17}. Methods in the other group focus on
making decisions more transparent, e.g., describing planning
decisions~\cite{borgo:ijcaiwrkshp18}, using partial order causal links
for explanations~\cite{seegebarth:icaps12}, combining classical first
order logic-based reasoning with interface design to help humans
understand a plan~\cite{bercher2014}, or using rules associated with
monotonic operators to define \emph{proof trees} that provide a
declarative view (i.e., explanation) of a
computation~\cite{ferrand:CI06}. There has also been work on
describing why a particular solution was obtained for a given problem
using non-monotonic logical reasoning~\cite{fandinno:TPLP19}.  These
methods are often agnostic to how an explanation is structured or
assume comprehensive domain knowledge. Methods are also being
developed to make the operation of deep networks more interpretable,
e.g., by computing gradients and constructing \emph{heat maps} of
relevant features~\cite{assaf:ijcai19,samek:ITU17}, or in the context
of deep networks trained to answer questions about images of
scenes~\cite{yi:nips18}.


Our work focuses on integrated robot systems that use a combination of
knowledge-based and data-driven algorithms to represent, reason with,
and learn from incomplete commonsense domain knowledge and noisy
observations. We seek to enable such robots to generate relational
descriptions of decisions, beliefs, and hypothetical or counterfactual
situations.  Recent surveys indicate that these capabilities are not
supported by existing systems~\cite{anjomshoae:aamas19,miller:AIJ19}.
Our architecture builds on existing work on making decisions more
transparent, and on work in our group on explainable
agency~\cite{langley:iaai17}, a theory of
explanations~\cite{mohan:KI19}, and on combining non-monotonic logical
reasoning and deep learning for scene understanding~\cite{mota:rss19}.

\section{Architecture}
\label{sec:arch}
\begin{figure}[tb]
  \centering
  \includegraphics[width=0.45\textwidth]{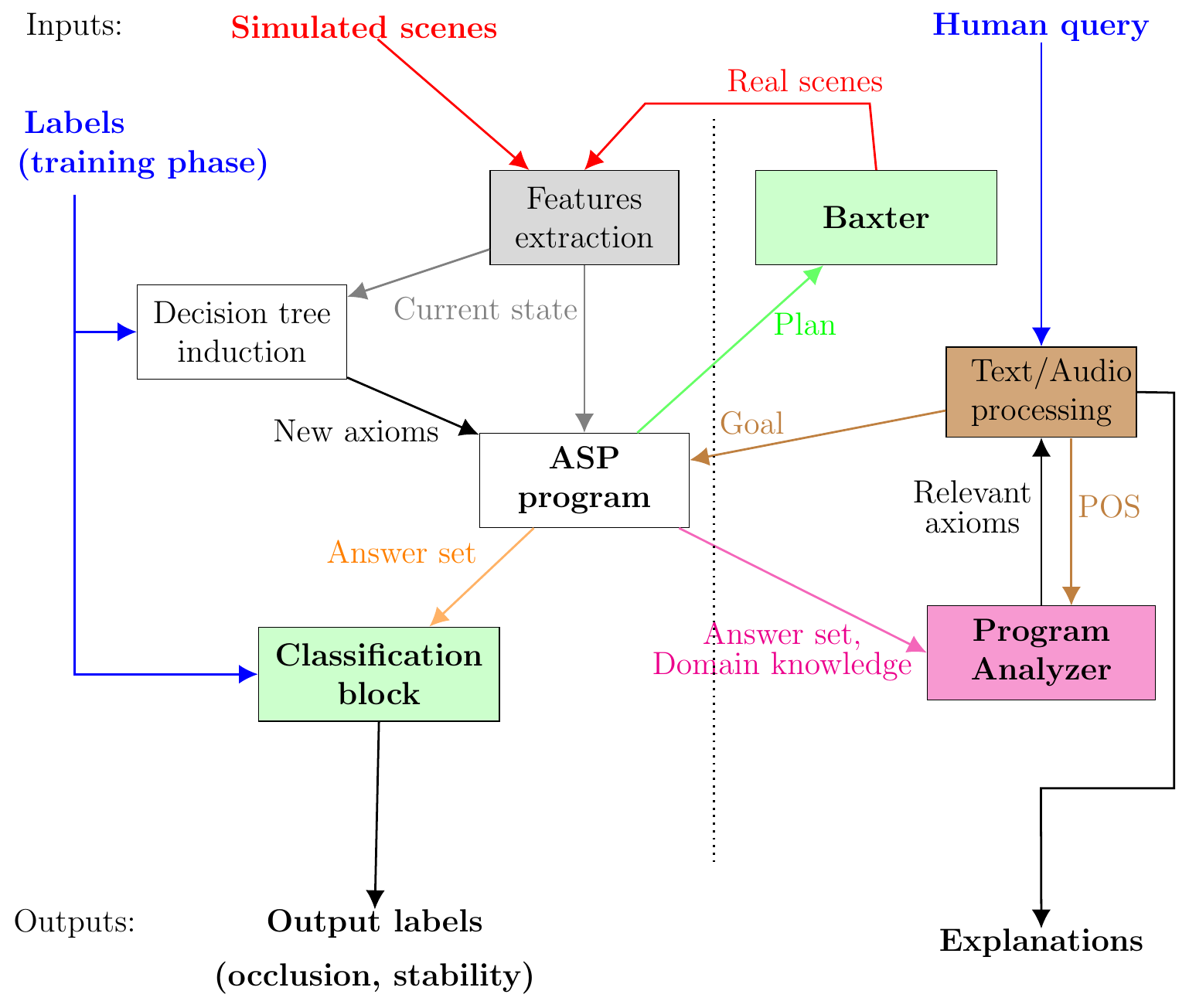}
  \vspace{-1em}
  \caption{Architecture combines non-monotonic logical reasoning, deep
    learning, and decision-tree induction. Components to the right of
    the dashed line support explainability. }
  \label{fig:architecture}
  \vspace{-1em}
\end{figure}

Figure~\ref{fig:architecture} shows the overall architecture.
Components to the left of the dashed vertical line combine
non-monotonic logical reasoning, deep learning, and decision-tree
induction for classification in simulated images~\cite{mota:rss19}.
Components to the right of the dashed line expand reasoning to explain
decisions, beliefs, and hypothetical situations~\cite{mota:eumas20}.
This paper extends the reasoning and learning capabilities to: (a)
learn action preconditions and effects from experience; and (b) trace
any given belief's evolution from the initial beliefs through the
application of specific axioms. We focus on the new components but
summarize all components for completeness, using the following example
domain.

\begin{example2}\label{ex:illus-example}[Robot Assistant (RA) Domain]\\
  {\rm A robot: (i) estimates occlusion of scene objects and stability
    of object structures, and arranges objects in desired
    configurations; and (ii) provides on-demand relational
    descriptions of decisions, beliefs, and hypothetical situations.
    There is uncertainty in the robot's perception and actuation, and
    probabilistic algorithms are used to visually recognize and move
    objects. The robot has incomplete domain knowledge, which includes
    object attributes such as $size$ (small, medium, large), $surface$
    (flat, irregular) and $shape$ (cube, apple, duck); spatial
    relations between objects (above, below, front, behind, right,
    left, in); some domain attributes; and some axioms governing
    domain dynamics such as:
    \begin{s_itemize}
    \item Placing an object on top of an object with an irregular
      surface results in an unstable object configuration.
    \item For any given object, removing all objects blocking the view
      of its frontal face causes it to be not occluded.
    \item An object below another object cannot be picked up.
    \end{s_itemize}
    This knowledge may need to be revised over time, e.g., some
    actions, axioms, and values of some attributes may be unknown, or
    the robot may find that placing certain objects on an object with
    an irregular surface does not cause instability.  }
\end{example2}

\subsection{Representation, Reasoning, Learning}
We first describe the knowledge representation, reasoning, and learning
components.

\subsubsection{Non-monotonic logical reasoning}
To represent and reason with domain knowledge, we use CR-Prolog, an
extension to Answer Set Prolog (ASP) that introduces \emph{consistency
  restoring} (CR) rules; we use the terms ``CR-Prolog'' and ``ASP''
interchangeably. ASP is a declarative language that represents
recursive definitions, defaults, causal relations, and constructs that
are difficult to express in classical logic formalisms. ASP is based
on the stable model semantics, and encodes \emph{default negation} and
\emph{epistemic disjunction}, e.g., unlike ``\stt{$\lnot$a}'', which
implies that ``\emph{a is believed to be false}'', ``\stt{not a}''
only implies ``\emph{a is not believed to be true}''. Each literal can
hence be true, false, or unknown. ASP supports non-monotonic logical
reasoning, i.e., adding a statement can reduce the set of inferences,
which helps recover from errors due to reasoning with incomplete
knowledge.

A domain's description in ASP comprises a \emph{system description}
$\mathcal{D}$ and a \emph{history} $\mathcal{H}$. $\mathcal{D}$
comprises a \emph{sorted signature} $\Sigma$ and axioms encoding the
domain's dynamics. In our prior work that explored spatial relations
for classification tasks, $\Sigma$ included \emph{basic sorts}, e.g.,
$object$, $robot$, $size$, $relation$, and $surface$; \emph{statics},
i.e., domain attributes that do not change over time, e.g.,
\textit{obj\_size(object, size)} and \textit{obj\_surface(obj,
  surface)}; and \emph{fluents}, i.e., attributes whose values can be
changed, e.g., \textit{obj\_rel(above, A, B)} implies object $A$ is
\emph{above} object $B$. Since the robot in this paper also plans and
executes physical actions that cause domain changes, we first describe
the expanded $\Sigma$ and transition diagram in action language
$\mathcal{AL}_d$~\cite{gelfond:ANCL13}, and then translate this
description to ASP statements. For the RA domain, $\Sigma$ now
includes the sort $step$ for temporal reasoning, additional fluents
such as \textit{in\_hand(robot, object)}, actions such as
\textit{pickup(robot, object)} and \textit{putdown(robot, object,
  location)}, and the relation \textit{holds(fluent, step)} implying
that a particular fluent holds true at a particular timestep.  Axioms
of the RA domain include $\mathcal{AL}_d$ statements such as:
\begin{align}
  \label{eqn:axiom-meta}
  \nonumber &putdown(rob_1, Ob_1, Ob_2)~~\mathbf{causes}~ obj\_rel(on, Ob_1, Ob_2)\\
  &obj\_rel(above, A, B)~~\mathbf{if}~obj\_rel(below, B, A) \\ \nonumber
  &\mathbf{impossible}~~pickup(rob_1, Ob_1)~\mathbf{if}~obj\_rel(below, Ob_1, Ob_2)
\end{align}
which encode a causal law, a state constraint, and an executability
condition respectively. Also, the domain's history $\mathcal{H}$
comprises records of fluents observed to be true or false, and of the
execution of an action, at a particular time step. We also expand
history to include initial state defaults.

The domain description is translated automatically to a CR-Prolog
program $\Pi(\mathcal{D}, \mathcal{H})$, which includes $\Sigma$ and
axioms of $\mathcal{D}$, inertia axioms, reality checks, closed world
assumptions for actions, and observations, actions, and defaults (with
CR rules) from $\mathcal{H}$; the program for the RA domain is
available online~\cite{code-results}. Planning, diagnostics, and
inference can then be reduced to computing \emph{answer sets} of
$\Pi$~\cite{gelfond:aibook14}. Any answer set represents the beliefs
describing a possible world; the literals of fluents and statics at a
time step represent the corresponding \emph{state}. Non-monotonic
logical reasoning allows the robot to recover from incorrect
inferences drawn due to incomplete knowledge, noisy sensors, or a low
threshold for elevating probabilistic information to logic statements.

\subsubsection{Classification:} For any given image, the robot tries
to estimate the occlusion of objects and the stability of object
configurations using ASP-based reasoning. If an answer is not found,
or an incorrect answer is found (on labeled training examples), the
robot automatically extracts relevant regions of interest (ROIs) from
the corresponding image. Parameters of Convolutional Neural Network
(CNN) architectures (Lenet~\cite{lecun:IEEE98},
AlexNet~\cite{krizhevsky:nips12}) are tuned to map information from
each such ROI to the corresponding classification labels.

\subsubsection{Decision tree induction:} Images used to train the CNNs
are considered to contain previously unknown information related to
occlusion and stability. Image features and spatial relations
extracted from ROIs in each such image, along with the known labels
for occlusion and stability (during training), are used to
learn a decision tree summarizing the corresponding state transitions. 
Next, branches of the tree that satisfy minimal thresholds on purity
at the leaf and have sufficient support from labeled examples are used
to construct candidate axioms.  Candidates are validated and those
without a minimal level of support on unseen examples are removed.
Also, we use an ensemble learning approach, retaining only axioms that
are identified over a number of cycles of learning and validation, and
axioms are merged to remove over-specifications.
In addition, each axiom is associated with a \emph{strength} that
decays exponentially over time if the axiom is not used or learned
again. Any axiom whose strength falls below a threshold is removed.

Our previous work only learned state constraints. In this paper, the
robot also learns previously unknown causal laws and executability
conditions if there is any mismatch between the expected and observed
state after an action is executed. Any expected but unobserved fluent
literal indicates missing executability condition(s), and any observed
unexpected fluent literal suggests missing causal law(s).
\begin{enumerate}
\item To explore missing executability conditions, the robot simulates
  the execution of the action (that caused the inconsistency) in
  different initial states and stores the relevant information from
  the initial state, executed action, and a label indicating the
  presence or absence of inconsistency. Any fluent literal in the
  answer set or initial state containing an object constant that
  occurs in the action, with variables replacing ground terms, is
  relevant.
    
\item To explore a missing causal law, training samples are collected
  as in Step 1, but the robot label is the unexpected fluent literal
  from the resultant state.

\item Separate decision trees are created with the relevant
  information from the initial state as the features (i.e., nodes) and
  the output labels (presence/absence of inconsistency for
  executability condition, unexpected fluent for causal law).  The
  root is the executed action.
\end{enumerate} 
Axioms are constructed from the decision trees as before.

\subsection{Relational Description as Explanation}
\label{sec:arch-explain}
The interplay between representation, reasoning, and learning is used
to provide relational descriptions of decisions, beliefs, and the
outcomes of hypothetical events.

\subsubsection{Interaction interface and control loop}
Existing software and a controlled (domain-specific) vocabulary are
used to parse human verbal (or text) input and to provide a response
when appropriate. Verbal input from a human is transcribed into text
based on the controlled vocabulary. This (or the input) text is
labeled using a part-of-speech (POS) tagger, and normalized with the
lemma list~\cite{someya1998} and related synonyms and antonyms from
WordNet~\cite{miller1995}. The processed text helps identify the type
of a desired goal or a request for information. Any given goal is sent
to the ASP program for planning, with the robot executing the plan
(and replanning when needed) until the goal is achieved. To address a
request for information, the ``Program Analyzer'' identifies the
relevant axioms and literals in the existing knowledge and inferred
beliefs. These literals are inserted into generic response templates
based on the controlled vocabulary, to provide textual or verbal
responses.

\subsubsection{Beliefs tracing}
\begin{figure*}[tb]
  \centering
      \includegraphics[width=0.7\textwidth]{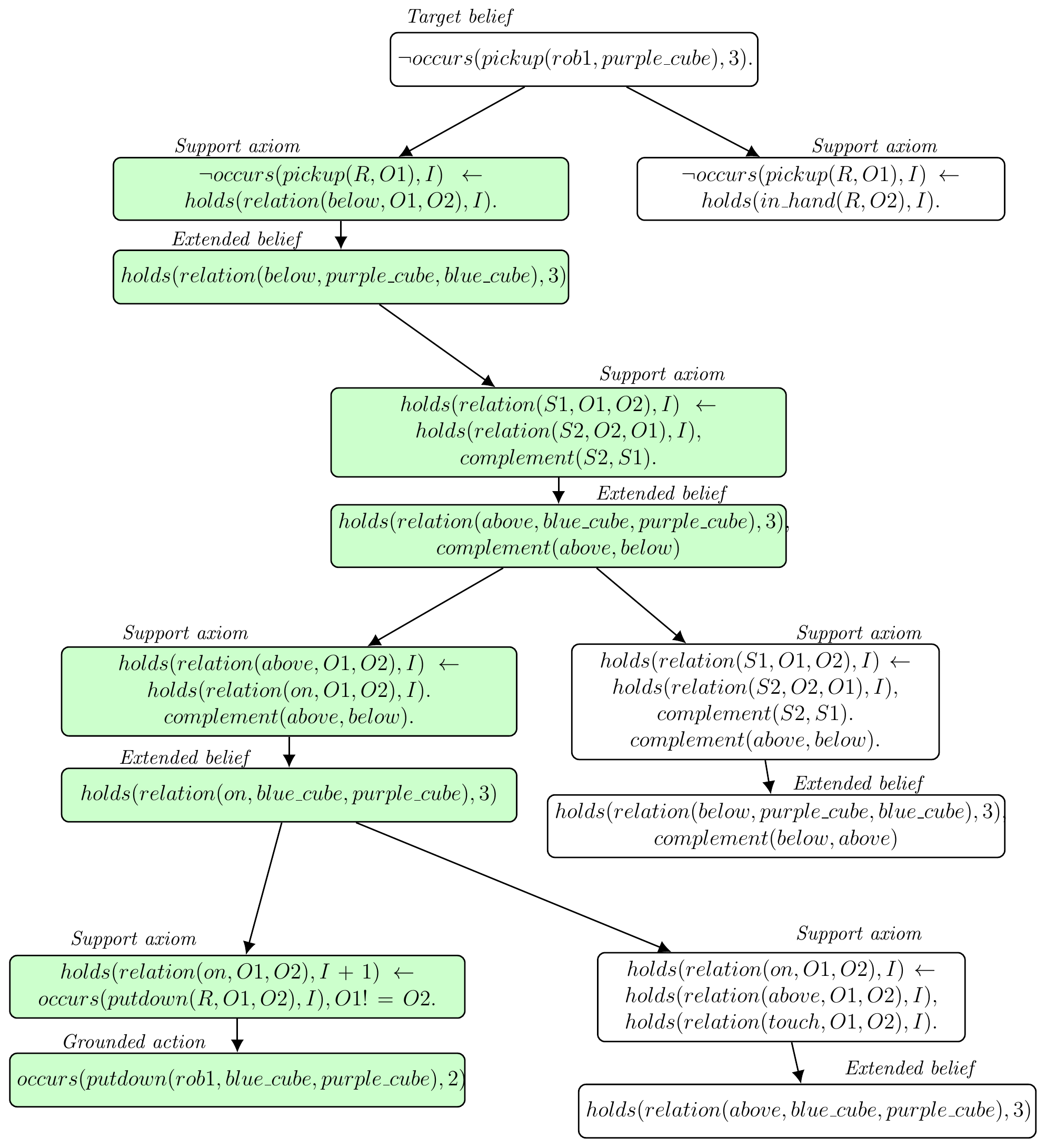}
  \caption{Example of belief tracing in action.}
  \label{fig:belief-tree}
\end{figure*}

A key capability of our architecture is to infer the sequence of
axioms whose application explains the evolution of any given belief.
Our approach adapts prior work on constructing such ``proof trees'',
which used monotonic logic statements to explain
observations~\cite{ferrand:CI06,genesereth:aibook87}, to our
non-monotonic logic formulation and traces the evolution of beliefs
corresponding to fluents or actions.
\begin{enumerate}
\item Select axioms whose head matches the belief of interest.
       
\item Ground the literals in the body of each selected axiom and check
  whether these are supported by the answer set.

\item Create a new branch in a proof tree (with target belief as root)
  for each selected axiom supported by the answer set, and store the
  axiom and the related supporting ground literals in suitable nodes.
    
\item Repeats Steps 1-3 with the supporting ground literals in Step 3
  as target beliefs in Step 1, until all branches reach a leaf node
  with no further supporting axioms.
\end{enumerate}
The paths from the root to the leaves in these proof trees help
construct the desired explanations. As an example, for the initial scenario in Figure~\ref{fig:left}, if the goal is to place the red cube on the orange cube, and the robot is asked (after plan execution) why it did not pick up the purple cube at time step 3, the corresponding proof tree would be as shown in Figure~\ref{fig:belief-tree}; the path highlighted in green contains the information needed to answer the question.

\subsubsection{Program analyzer}
We illustrate our approach for constructing explanations (in the form
of relational descriptions) in the context of four types of
\emph{explanatory} questions or requests. The first three were
introduced as question types to be considered by any explainable
planning system~\cite{fox:ijcaiwrkshp17}; we also consider a question
about the robot's beliefs at any point in time.
\begin{enumerate}
\item \underline{\textbf{Plan description}} When asked to describe a
  plan, the robot parses the related answer set(s) and extract a
  sequence of actions such as \textit{occurs(action1, step1), ...,
    occurs(actionN, stepN)} to construct the response.

\vspace{0.5em}
\item \underline{\textbf{Action justification: Why action X at step
      I?}}~ To justify the execution of an action at a particular time
  step:
  \begin{enumerate}
  \item For each action that occurred after time step $I$, the robot
    examines relevant executability condition(s) and identifies
    literal(s) that would prevent the action's execution at step $I$.
    For the goal of picking up the $orange\_block$ in
    Figure~\ref{fig:left}, assume that the executed actions are
    \textit{occurs(pickup(robot, blue\_block), 0)},
    \textit{occurs(putdown(robot, blue\_block), 1)}, and
    \textit{occurs(pickup(robot, orange\_block), 2)}. If the focus is
    on the first $pickup$ action, an executability condition related
    to the second $pickup$ action:
    \begin{align*}
      \neg occurs&(pickup(robot, A), I) ~\leftarrow\\ \nonumber
      &holds(obj\_rel(below, A, B), I)
    \end{align*}
    is ground in the scene to obtain \textit{obj\_rel(below,\
      orange\_block,\ blue\_block)} as a literal of interest.
    
  \item If any identified literal is in the answer set at the time
    step of interest ($0$ in this example) and is absent (or its
    negation is present) in the next step, it is a reason for
    executing the action under consideration.
    
  \item The condition modified by the execution of the action of
    interest is paired with the subsequent action to construct the
    answer to the question. The question ``Why did you pick up the
    blue block at time step 0?'', receives the answer ``I had to pick
    up the orange block, and the orange block was below the blue
    block''.
  \end{enumerate} 
  A similar approach is used to justify the selection of any
  particular action in a plan that has not been executed.

  \vspace{0.5em} 
\item \underline{\textbf{Hypothetical actions: Why not action X at
      step I?}}~ For questions about actions not selected for
  execution:
  \begin{enumerate} 
  \item The robot identifies executability conditions that have the
    hypothetical action in the head, i.e., conditions that prevent the
    action from being selected during planning.

  \item For each such executability condition, the robot checks if
    literals in the body are satisfied by the corresponding answer
    set. If yes, these literals form the answer.
  \end{enumerate} 
  Suppose action \textit{putdown(robot,\ blue\_block,\ table)}
  occurred at step $1$ in Figure~\ref{fig:left}. For the question
  ``Why did you not put the blue cube on the tennis ball at time step
  1?'', the following executability condition is identified:
  \begin{align*}
    \neg occurs(putdown&(robot,\ A,\ B), I) ~\leftarrow\\\nonumber
    &has\_surface(B,\ irregular)
  \end{align*}
  which implies that an object cannot be placed on another object with
  an irregular surface. The answer set states that the tennis ball has
  an irregular surface and the robot answers ``Because the tennis ball
  has an irregular surface''.  This process uses the \emph{belief
    tracing} approach.

\vspace{0.5em}
\item \underline{\textbf{Belief query: Why belief Y at step I?}}~ To
  explain any particular belief, the robot uses the \emph{belief
    tracing} approach described earlier. The supporting axioms and
  relevant literals identified are used to construct the answer.  For
  instance, to explain the belief that object $ob_1$ is unstable in
  step $I$, the robot finds the support axiom:
  \begin{align*}
    \nonumber \neg holds(stable(ob_1), I) ~\leftarrow~
    holds(small\_base(ob_1), I)
  \end{align*}   
  Assume that the current beliefs include that $ob_1$ has a small
  base.  Tracing this belief identifies the axiom:
  \begin{align*}
    \nonumber
    holds&(small\_base(ob_1), I) ~\leftarrow\\\nonumber &holds(relation(below,\ ob_2,\ ob_1),\ I),\\
    \nonumber &has\_size(ob_2,\ small),\ has\_size(ob_1,\ big)
  \end{align*}
  Asking ``why do you believe object $ob_1$ is unstable at step I?''
  would provide the answer ``Because object $ob_2$ is below object
  $ob_1$, $ob_2$ is small, and $ob_1$ is big''.
\end{enumerate}

\subsubsection{Robot platform}
As stated earlier, our work consider scene understanding tasks and
planning tasks. For robot experiments, we use a Baxter manipulating
objects on a tabletop. The Baxter uses probabilistic algorithms to
process inputs from its cameras, e.g., to detect objects, their
attributes, and the spatial relations between them, from images. It
also uses probabilistic motion planning algorithms to execute
primitive manipulation actions, e.g., to grasp and pick up objects.
Observations obtained with a high probability are elevated to literals
with complete certainty in the ASP program.


\section{Experimental Setup and Results}
\label{sec:expres}
We present execution traces and quantitative results illustrating the
ability to construct relational descriptions of decisions, beliefs,
and hypothetical events; and to learn causal laws and executability
conditions.

\subsection{Experimental Setup}
\label{sec:expres-setup}
We experimentally evaluated the following hypotheses:
\begin{itemize}
\item[\underline{\textbf{H1}}]: our architecture enables the robot to
  accurately learn previously unknown domain axioms;

\item[\underline{\textbf{H2}}]: reasoning with incrementally learned
  axioms improves the quality of plans generated;

\item[\underline{\textbf{H3}}]: the beliefs tracing approach
  accurately retrieves the supporting axioms associated with any
  belief; and

\item[\underline{\textbf{H4}}]: exploiting the links between reasoning
  and learning improves the accuracy of the explanatory descriptions.
\end{itemize}
These hypotheses and our architecture's capabilities were evaluated in
the context of the four types of requests described earlier, but the
methodology can be adapted for other types of requests.  Plan quality
was measured in terms of the ability to compute minimal and correct
plans. The quality of an explanation was measured in terms of
precision and recall of its literals in comparison with the expected
(``ground truth'') response obtained in a semi-supervised manner based
on manual input and automatically selected relevant literals.

Experimental trials considered images from the robot's camera and
simulated images. Real world images contained $5-7$ objects of
different colors, textures, shapes, and sizes in the RA domain
(Example~\ref{ex:illus-example}). The objects included cubes, a pig, a
capsicum, a tennis ball, an apple, an orange, and a pot. These objects
were either stacked on each other or spread on the table---see
Figure~\ref{fig:left}.  A total of $20$ configurations were created,
each with five different goals for planning and four different
questions for each plan, resulting in $100$ plans and $400$ questions.
Since it is time-consuming and difficult to run many trials on robots,
we also used a real-time physics engine (Bullet) to create $20$
simulated images, each with $7-9$ objects ($3-5$ stacked and the
remaining on a flat surface).  Objects included cylinders, spheres,
cubes, a duck, and five household objects from the Yale-CMU-Berkeley
dataset (apple, pitcher, mustard bottle, mug, and box of crackers). We
once again considered five different goals for planning and four
different questions for each plan, resulting in (once again) $100$
plans and $400$ questions.

To explore the interplay between reasoning and learning, we focused on
the effect of learned knowledge on planning and constructing
explanations. We ran experiments with and without some learned axioms
in the knowledge base. Learned axioms were revised over time in our
architecture, whereas these axioms were not used by the baselines for
planning and explanation generation. During planning, we measured the
number of optimal, sub-optimal, and incorrect plans, and the planning
time. An \emph{optimal} plan is a minimal plan that achieves the goal;
a \emph{sub-optimal} plan requires more than the minimum number of
steps and/or has to assume an unnecessary exception to defaults; and
an \emph{incorrect} plan leads to undesirable outcomes and fails to
achieve the goal.

To test hypothesis \textbf{\underline{H1}} we removed five axioms
(three executability conditions and two causal laws) from the agent's
knowledge, and ran the learning algorithm $20$ times. The robot
executed actions to learn all the missing axioms each time.  Each run
stops if the robot executes a number of actions without detecting any
inconsistency, or if a maximum number of decision trees are
constructed.  The overall precision and recall are then computed.

\subsection{Execution Traces}
\label{sec:expres-trace}
The following execution traces illustrate our approach to construct
relational descriptions explaining the decisions, beliefs, and the
outcomes of hypothetical actions.

\begin{figure}[tb]
  \centering
      \includegraphics[width=0.35\textwidth]{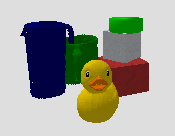}
  \caption{Simulated scene for the execution examples.}
  \label{fig:exec-examples}
\end{figure}

\begin{execexample}\label{exec-example1}[Plans, actions, and beliefs] \\
  {\rm Consider a scene with objects as shown in
    Figure~\ref{fig:exec-examples}. The robot's goal is to achieve a
    state in which the pitcher is on the red block, i.e.,
    \textit{holds(relation(on, pitcher, red\_block), I)}.  The robot
    answers the following questions \emph{after executing} a plan and
    successfully achieving the assigned goal:
    \begin{itemize}
    \item \textbf{Human:} ``Please describe the plan.''
      
      \textbf{Baxter:} ``I picked up the green can. I put the green
      can on the table. I picked up the white block. I put the white
      block on the green can. I picked up the pitcher. I put the
      pitcher on the red block.''
      
      \smallskip
    \item The human may ask the robot to justify a particular action.
      
      \textbf{Human:} ``Why did you pick up the green can at step 0?''
      
      \textbf{Baxter:} ``Because I had to pick up the white block, and
      it was below the green can.''

      \smallskip
    \item The human may ask about actions not chosen.
  
      \textbf{Human:} ``Why did you not put white block on the mug?''
  
      \textbf{Baxter:} ``Because the mug has irregular surface.''
      Since there was no reference to a particular time step, the
      robot responds based on the single instance (in the executed
      plan) of putting the white block on the mug.

      \smallskip
    \item The human may also ask about particular beliefs.
  
      \textbf{Human:} ``Why did you believe that the white block was
      below the green can in the initial state?''

      \textbf{Baxter:} ``Because I observed the white block below the
      green can at step zero.''

    \end{itemize}
  }
\end{execexample}

\begin{execexample}\label{exec-example2}[Beliefs tracing and explanation] \\
  {\rm We continue with our previous example:
    \begin{itemize}
    \item \textbf{Human:} ``Why did you not pick white block at step
      0?''
    
      The robot uses the belief tracing approach to construct a proof
      tree with \textit{$\neg$ occurs(pickup(rob1, white\_block), 0)}
      as the root. For each axiom in which this ground literal matches
      the head, it checks if its body is supported by the answer set.
      If yes, ground literals in the body are used to expand the tree.
      According to the third axiom in Equation~\ref{eqn:axiom-meta},
      one of the extended beliefs is \textit{holds(obj\_rel(below,
        white\_block, green\_can), 0)}.  Similar searches are repeated
      until no further supporting axioms are found. In our example,
      the statement \textit{holds(relation(on, white\_block,
        green\_can), 0)} is output as the leaf of the proof tree, and
      the agent's answer to the question is:
    
    \item \textbf{Robot:} ``Because I observed the green can on the
      white block at step 0.''

    \end{itemize}
  }       
\end{execexample}

\begin{execexample}\label{exec-example3}[Learning and explanation] \\
  {\rm In some situations, the robot may not possess the knowledge
    required to address the human request. Continuing with the
    previous example, the human may ask:
    \begin{itemize}

    \item \textbf{Human:} ``Why did you not pick up green can at step
      5?''

      By creating a proof tree, the answer is found:\\
      \textbf{Robot:} ``Because white block was on the green can.''
      
      The human may need further details and ask:

      \textbf{Human:} ``Why did you believe the white block was on the
      green can?''
      
      To answer this question the robot has to know the causal
      relation between action \textit{putdown} and the spatial
      relation \textit{on}---first axiom in
      Equation~\ref{eqn:axiom-meta}. After the robot learns this
      causal law, it produces the correct answer:

      \textbf{Robot:} ``Because I put the white block on the green can
      at step 4.''
    \end{itemize}
    This example illustrates the benefit of integrating reasoning and
    learning to justify particular beliefs. }
\end{execexample}
\noindent
Overall, these (and other) examples show the ability to focus on
relevant knowledge, incrementally revise axioms, trace relevant
beliefs, and identify attributes and actions relevant to a given
scenario. They also support hypothesis \textbf{\underline{H3}}.
 

\subsection{Experimental Results}
\label{sec:expres-results}
The first set of experiments evaluated \textbf{\underline{H1}}. We
removed five axioms (two causal laws and three executability
conditions) from the robot's knowledge, and ran the learning algorithm
$20$ times. We measured the precision and recall for the missing
axioms in each run, and table \ref{tab:axioms} summarizes the results.
The row labeled ``Strict'' provides results when any variation in the
target axiom is considered an error. In this case, even over-specified
axioms, i.e., axioms that have some additional irrelevant literals,
are considered to be incorrect. Equation~\ref{eq.:causalover} shows
one example of such an axiom in which the second literal in the body
is irrelevant. The row labeled ``Relaxed'' reports results when
over-specifications are not considered errors; the high precision and
recall support \textbf{\underline{H1}}.

\begin{table}[htb]
\caption{Precision and recall for learning previously unknown axioms using decision tree induction.}
\centering
\begin{tabular}{| c | c | c|}\hline
\textbf{Missing Axioms} & \textbf{Precision} & \textbf{Recall}\\[0.0ex]
\hline
&\\[-2.4ex]
Strict & 69.2\% & 78.3\%\\[0.0ex]
\hline
&\\[-2.4ex]
Relaxed & 96\%  & 95.1\%\\[0.0ex] \hline
\end{tabular}
\vspace{-1.75em}
\label{tab:axioms}
\end{table}

\begin{align}
 \nonumber \neg holds&(in\_hand(R1, O1),I+1)~\leftarrow\\\nonumber
   &occurs(putdown(R1, O1, O2),I),\\
   &\neg holds(in\_hand(R1,  O5),I ). 
\label{eq.:causalover}
\end{align}

\noindent
The second set of experiments was designed to evaluate hypothesis
\textbf{\underline{H2}}.
\begin{enumerate}
\item As stated earlier, $20$ initial object configurations were
  created (similar to Figure~\ref{fig:initial}). The Baxter
  automatically extracted information (e.g., attributes, spatial
  relations) from images corresponding to top and frontal views
  (cameras on the left and right grippers), and encoded it in the ASP
  program as the initial state.

\item For each initial state, five goals were randomly chosen and
  encoded in the ASP program. The robot reasoned with the existing
  knowledge to create plans for these $100$ combinations ($20$ initial
  states, five goals).
  
\item The plans were evaluated in terms of the number of optimal,
  sub-optimal and incorrect plans, and planning time.

\item Trials were repeated with and without learned axioms, and for
  the simulated images.
\end{enumerate}
Since the number of plans and planning time vary depending on the initial conditions and the goal, we conducted paired trials with and without the learned axioms included in the ASP program used for
reasoning. The initial conditions and goal were identical in each paired trial, but differed between paired trials. Then, we expressed the number of plans and the planning time with the learned axioms as a fraction of the corresponding values obtained by reasoning without the learned axioms. The average of these fractions over all the trials is reported in Table~\ref{tab:plans}. We also computed the number of optimal, sub-optimal, and incorrect plans in each trial as a fraction of the total number of plans; we did this with and without using the learned axioms for reasoning, and the average over all trials is
summarized in Table~\ref{tab:plans2}. 

These results indicate that for images of real scenes, using the
learned axioms for reasoning significantly reduced the search space,
resulting in a much smaller number of plans and a substantial
reduction in the planning time. The use of the learned axioms does not
seem to make any significant difference with the simulated scenes.
This is understandable because simulated images have more objects with
several of them being small objects. This increases the number of
possible plans to achieve any given goal. In addition, when the robot
used the learned axioms for reasoning, it reduced the number of
sub-optimal plans and eliminated all incorrect plans. Also, almost
every sub-optimal plan was created when the corresponding goal could
not be achieved without creating an exception to a default. Without
the learned axioms, a larger fraction of the plans are sub-optimal or
incorrect. Note that the number of suboptimal plans is higher with
simulated scenes that have more objects to consider. These results
support hypothesis \textbf{\underline{H2}} but also indicate the need
to explore complex scenes further.

\begin{table}[tb]
  \caption{Number of plans and planning time with the learned axioms expressed as a fraction of the values without the learned axioms.}
  \label{tab:plans}
  \vspace{-1em}
  \begin{center}
    \begin{tabular}{|c|c|c|}
      \hline
      & \multicolumn{2}{|c|}{\textbf{Ratio (with/without)}}\\      
      \cline{2-3}
      \textbf{Measures} & \textbf{Real scenes} & \textbf{Simulated scenes}\\
      \hline
      Number of steps & 1.17 & 1.21\\
      \hline
      Number of plans & 0.7 & 1.1\\
      \hline
      Planning time & 0.87 & 1.08\\
      \hline
    \end{tabular}
  \end{center}
\end{table}

\begin{table}[tb]
  \caption{Number of optimal, sub-optimal, and incorrect plans expressed as a fraction of the total number of plans. Reasoning with the learned axioms improves performance.}
  \label{tab:plans2}
  \vspace{-1em}
  \begin{center}
    \begin{tabular}{|c|c|c|c|c|}
      \hline
      & \multicolumn{2}{|c|}{\bf Real Scenes} & \multicolumn{2}{|c|}{\bf Simulated Scenes} \\ \cline{2-5}
      \textbf{Plans} & \textbf{Without} & \textbf{With} & \textbf{Without} & \textbf{With}\\
      \hline
      Optimal & 0.33 & 0.89 & 0.13 & 0.24 \\
      \hline
      Sub-optimal & 0.12 & 0.11 & 0.44 & 0.76 \\
      \hline
      Incorrect & 0.55 & 0 & 0.43 & 0 \\
      \hline
    \end{tabular}
  \end{center}
  \vspace{-1.5em}
\end{table}

\medskip
\noindent
The third set of experiments was designed as follows to evaluate
hypothesis \textbf{\underline{H4}}:
\begin{enumerate}
\item For each of the $100$ combinations ($20$ configurations, five
  goals) from the first set of experiments with real-world data, we
  considered knowledge bases with and without the learned axioms and
  had the robot compute plans to achieve the goals.

\item The robot had to describe the plan and justify the choice of a
  particular action (chosen randomly) in the plan. Then, one parameter
  of the chosen action was changed randomly to pose a question about
  why this new action could not be applied. Finally, a belief related
  to the previous two questions had to be justified.

\item The literals present in the answers were compared against the
  expected literals in the ``ground truth'' response, with the average
  precision and recall scores reported in
  Table~\ref{tab:explanations1}.

\item We also performed these experiments with simulated images, and the results are summarized in Table~\ref{tab:explanations2}.
\end{enumerate}

\begin{table}[t]
  \caption{(\textbf{Real scenes}) Precision and recall of retrieving relevant literals for constructing answers to questions with and without using the learned axioms for reasoning. Using the learned axioms significantly improves the ability to provide accurate explanations.}
  \label{tab:explanations1}
  \vspace{-1em}
  \begin{center}
    \scalebox{0.93}{%
      \begin{tabular}{|c|c|c|c|c|}
        \hline
        & \multicolumn{2}{|c|}{\bf Precision} & \multicolumn{2}{|c|}{\bf Recall} \\ \cline{2-5}
        \hline
        \textbf{Query Type} & \textbf{Without} & \textbf{With} & \textbf{Without} & \textbf{With}\\
        \hline
        Plan description & 74.94\% & 100\% & 63.25\% & 100\%\\
        \hline
        Why X? & 72.22\% & 94.0\% & 65.0\% & 94.0\%\\
        \hline
        Why not X? & 100\% & 95.92\% & 68.89\% & 100\%\\
        \hline
        Belief & 95.74\% & 100\% & 95.74\% & 100\%\\
        \hline
      \end{tabular}}
  \end{center}
\end{table}

\begin{table}[t]
  \caption{(\textbf{Simulated scenes}) Precision and recall of retrieving relevant literals for constructing answers to questions with and without reasoning with learned axioms. Using the learned axioms significantly improves the ability to provide accurate explanations.}
  \label{tab:explanations2}
  \vspace{-1em}
  \begin{center}
    \scalebox{0.93}{%
      \begin{tabular}{|c|c|c|c|c|}
        \hline
        & \multicolumn{2}{|c|}{\bf Precision} & \multicolumn{2}{|c|}{\bf Recall} \\ \cline{2-5}
        \hline
        \textbf{Query Type} & \textbf{Without} & \textbf{With} & \textbf{Without} & \textbf{With}\\
        \hline
        Plan description & 71.85\% & 100\% & 59.39\% & 100\%\\
        \hline
        Why X? & 66.48\% & 95.0\% & 58.5\% & 95.0\%\\
        \hline
        Why not X? & 86.79\% & 95.24\% & 63.01\% & 100\%\\
        \hline
        Belief & 94.55\% & 100\% & 91.23\% & 100\%\\
        \hline
      \end{tabular}}
  \end{center}
\end{table}

\noindent
Tables~\ref{tab:explanations1},~\ref{tab:explanations2} show that when
the learned axioms were used for reasoning, the precision and recall
of relevant literals (for constructing the explanation) were higher
than when the learned axioms were not included. The improvement in
performance is particularly pronounced when the robot has to answer
questions about actions that it has not actually executed. The
precision and recall rates were reasonable even when the learned
axioms were not included; this is because not all the learned axioms
are needed to accurately answer each explanatory question. When the
learned axioms were used for reasoning, errors were very rare and
corresponded to some additional literals being included in the answer
(i.e., over-specified explanations). In addition, when we specifically
removed axioms related to the goal under consideration, precision and
recall values were much lower.  Furthermore, there was noise in both
sensing and actuation, especially in the robot experiments. For
instance, recognition of spatial relations, learning of constraints,
and manipulation have approximate error rates of $15\%$, $5-10\%$, and
$15\%$ respectively.  Experimental results thus indicate that coupling
reasoning and learning to inform and guide each other enables the
robot to provide accurate relational descriptions of decisions,
beliefs, and the outcomes of hypothetical actions. This supports
hypothesis \textbf{\underline{H4}}.  Additional examples of images,
questions, and answers, are in our open source
repository~\cite{code-results}.

\section{Conclusions}
\label{sec:conclusion}
This paper described an approach inspired by cognitive systems and knowledge representation tools to enable an integrated robot system to explain its decisions, beliefs, and the outcomes of hypothetical actions. These explanations are constructed on-demand in the form of descriptions of relations between
relevant objects, actions, and domain attributes. We have implemented this approach in an architecture that combines the complementary strengths of non-monotonic logical reasoning with incomplete commonsense domain knowledge, deep learning, and decision tree induction.  In the context of some scene understanding and planning tasks performed in simulation and a physical robot, we have demonstrated that our architecture exploits the interplay between knowledge-based reasoning and data-driven learning.  It automatically identifies and reasons with the relevant information to efficiently construct the desired explanations, with both the planning and explanation generation performance improving when previously unknown axioms are learned and used for subsequent reasoning.

Our architecture opens up multiple avenues for further research.
First, we will explore more complex domains, tasks, and explanations,
reasoning with relevant knowledge at different tightly-coupled
resolutions for scalability~\cite{mohan:JAIR19}. We are specifically
interested in exploring scenarios in which there is ambiguity in the
questions (e.g., it is unclear which of two occurrences of the
$pickup$ action the human is referring to), or the explanation is
needed at a different level of abstraction, specificity, or verbosity.
We will do so by building on a related theory of
explanations~\cite{mohan:KI19}. Second, we will use our architecture
to better understand the behavior of deep networks. The key advantage
of using our architecture is that it uses reasoning to guide learning.
Unlike ``end to end'' data-driven deep learning methods, our
architecture uses reasoning to trigger learning only when existing
knowledge is insufficient to perform the desired task(s). The
long-term objective is to develop an architecture that exploits the
complementary strengths of knowledge-based reasoning and data-driven
learning for the reliable and efficient operation of robots in
complex, dynamic domains.


\bibliographystyle{aaai} 


\end{document}